\documentclass{article}

\usepackage{microtype}
\usepackage{graphicx}
\usepackage{booktabs}
\usepackage[skip=6pt, font=small, labelfont=bf]{caption}
\usepackage{hyperref}

\usepackage[accepted]{icml2025}

\usepackage{amsmath}
\usepackage{amssymb}
\usepackage{multirow}
\usepackage{algorithm}
\usepackage{algorithmic}
\usepackage{wrapfig}
\usepackage{listings}
\usepackage{longtable}
\usepackage{enumitem}
\usepackage{tikz}
\usepackage{mathpazo}
\usepackage[capitalize,noabbrev]{cleveref}


\usepackage{fontspec}
\usepackage{polyglossia}
\setmainlanguage{english}
\setotherlanguage{amharic}
\newfontfamily\geezfont{GeezNotoSans-Light}[Path=./,Extension=.ttf,Script=Ethiopic]
\newfontfamily\geezfontbold{GeezNotoSans-SemiBold}[Path=./,Extension=.ttf,Script=Ethiopic]
\newcommand{\ti}[1]{{\geezfont #1}}
\newcommand{\tib}[1]{{\geezfontbold #1}}
\newcommand{\tiipa}[2]{{\geezfont #1} /#2/}

\definecolor{mydarkblue}{rgb}{0.0,0.15,0.7}
\hypersetup{colorlinks=true,
linkcolor=mydarkblue,
citecolor=mydarkblue,
filecolor=mydarkblue,
urlcolor=mydarkblue}

\icmltitlerunning{Tigrinya Number Verbalization}

\begin{document}

\icmltitle{Tigrinya Number Verbalization: \\
Rules, Algorithm, and Implementation}

\begin{icmlauthorlist}
\end{icmlauthorlist}

\begin{center}
\begin{minipage}[t]{0.25\textwidth}
\centering
{\bf Fitsum Gaim} \\
\texttt{fitsum@geezlab.com}
\end{minipage}
\hspace{0.1cm}
\begin{minipage}[t]{0.25\textwidth}
\centering
{\bf Issayas Tesfamariam} \\
\texttt{issayas@stanford.edu}
\end{minipage}
\\[1em]
{Preprint | Jan. 2026}
\end{center}

\vskip 0.3in

\begin{abstract}
We present a systematic formalization of Tigrinya cardinal and ordinal number verbalization, addressing a gap in computational resources for the language. This work documents the canonical rules governing the expression of numerical values in spoken Tigrinya, including the conjunction system, scale words, and special cases for dates, times, and currency. We provide a formal algorithm for number-to-word conversion and release an open-source implementation.\footnote{Tigrinya-Numbers package available at \url{https://github.com/fgaim/tigrinya-numbers}} Evaluation of frontier large language models (LLMs) reveals significant gaps in their ability to accurately verbalize Tigrinya numbers, underscoring the need for explicit rule documentation. This work serves language modeling, speech synthesis, and accessibility applications targeting Tigrinya-speaking communities.
\end{abstract}

\section{Introduction}

Tigrinya (\ti{ትግርኛ}) is a South Semitic language of the Afroasiatic family, primarily spoken in Eritrea and Ethiopia. With approximately 10 million speakers, it ranks among the most widely spoken Semitic languages globally \citep{eberhard-etal-2024-ethnologue}. Despite this, computational linguistic resources for Tigrinya remain limited \citep{gaim-park-2025-tigrinya-nlp}, even compared to related languages such as Amharic.
Number verbalization, the conversion of numerical digits to their spoken word forms, is a fundamental component of natural language processing pipelines. It serves as a preprocessing step for text-to-speech (TTS) synthesis, automatic speech recognition (ASR) language modeling, and accessibility technologies. While number verbalization systems exist for many languages through libraries such as \texttt{num2words} \citep{num2words}, Tigrinya is notably absent from such resources.

This technical note presents the canonical rules for Tigrinya number verbalization (covering cardinals, ordinals, dates, times, currency, and telephone numbers), presents a formal conversion algorithm, and an open-source implementation in Python. We also evaluate frontier large language models on this task to assess their current capabilities.

\section{The Tigrinya Number System}

Tigrinya employs a decimal number system with distinct lexical items for digits 0--9, multiples of ten (10--90), and scale words for powers of ten and beyond.\footnote{While Ge'ez numerals exist in historical and religious scripts, modern Tigrinya uses the Arabic numerals.} The verbalization follows compositional rules involving a conjunction suffix that marks compound structures. Table~\ref{tab:numbers} presents the canonical forms for digits, tens, scales, and ordinals, consistent with standard Tigrinya language pedagogy and instructions \citep{tesfamariam-2018-lets-speak-tigrinya}.

\subsection{Cardinal Numbers}
\label{sec:cardinals}

\textbf{Zero} is expressed as \tiipa{ዜሮ}{zero} a loanword or \tiipa{ባዶ}{bado} a native term. \textbf{Digits (1--10)} form the foundation of the Tigrinya number verbalization system (see Table~\ref{tab:numbers}). \textbf{Teens (11--19)} are formed by juxtaposing \tiipa{ዓሰርተ}{`asärtä} (ten) with the ones digit, separated by a space but without the conjunction suffix (e.g., 15 $\rightarrow$ \tiipa{ዓሰርተ ሓሙሽተ}{`asärtä ħamushtä}). \textbf{Tens (20--90)} have unique suppletive forms distinct from their corresponding digits.
Tigrinya uses loanwords for scale numbers beyond thousand (million, billion, trillion, etc). Negative numbers are prefixed with \tiipa{ኣሉታ}{'aluta}, decimals use \tiipa{ነጥቢ}{näTbi} (point) followed by the mantissa read digit-by-digit, and percentages append \tiipa{ሚእታዊት}{mi'tawit} (e.g., 40\% $\rightarrow$ \tiipa{ኣርብዓ ሚእታዊት}{'arb`a mi'tawit}).
Tigrinya cardinal numerals are generally gender-invariant, except `one' (\tiipa{ሓደ}{ħadä} vs. \tiipa{ሓንቲ}{ħanti}), which preserves morphological gender agreement in standalone contexts. In compound contexts (e.g., 21, 101), this distinction is neutralized, defaulting to the masculine form (\ti{ሓደ}) irrespective of the head noun's gender.

\subsection{Ordinal Numbers}

Tigrinya ordinal numbers 1st--10th have unique suppletive forms with gender distinction (masculine and feminine). For ordinals 11th and above, a prefix construction is used: \tiipa{መበል}{mäbäl} + cardinal number, so `25th' is read as \tiipa{መበል ዕስራን ሓሙሽተን}{mäbäl `sran ħamushtän}.

\begin{table}[t]
\caption{Tigrinya Cardinal and Ordinal Number Words. Ordinals are gendered as masculine and feminine.}
\label{tab:numbers}
\centering
\footnotesize
\begin{tabular}{@{\hspace{3pt}}cl@{\hspace{9pt}}cl@{\hspace{9pt}}ll@{\hspace{9pt}}ll@{\hspace{3pt}}}
\toprule
\multicolumn{2}{c}{\textbf{Digits}} & \multicolumn{2}{c}{\textbf{Tens}} & \multicolumn{2}{c}{\textbf{Scales}} & \multicolumn{2}{c}{\textbf{Ordinals} (Masc./Fem.)} \\
\cmidrule(r){1-2} \cmidrule(lr){3-4} \cmidrule(lr){5-6} \cmidrule(l){7-8}
1 & \tiipa{ሓደ}{ħadä} & 10 & \tiipa{ዓሰርተ}{`asärtä} & $10^2$ & \tiipa{ሚእቲ}{mi'ti} & 1st & \tiipa{ቀዳማይ}{qädamay} \tiipa{ቀዳመይቲ}{qädamäyti} \\
2 & \tiipa{ክልተ}{kltä} & 20 & \tiipa{ዕስራ}{`sra} & $10^3$ & \tiipa{ሽሕ}{shħ} & 2nd & \tiipa{ካልኣይ}{kal'ay} \tiipa{ካልአይቲ}{kal'äyti} \\
3 & \tiipa{ሰለስተ}{sälästä} & 30 & \tiipa{ሰላሳ}{sälasa} & $10^6$ & \tiipa{ሚልዮን}{milyon} & 3rd & \tiipa{ሳልሳይ}{salsay} \tiipa{ሳልሰይቲ}{salsäyti} \\
4 & \tiipa{ኣርባዕተ}{'arba`tä} & 40 & \tiipa{ኣርብዓ}{'arb`a} & $10^9$ & \tiipa{ቢልዮን}{bilyon} & 4th & \tiipa{ራብዓይ}{rab`ay} \tiipa{ራብዐይቲ}{rab`äyti} \\
5 & \tiipa{ሓሙሽተ}{ħamushtä} & 50 & \tiipa{ሓምሳ}{ħamsa} & $10^{12}$ & \tiipa{ትሪልዮን}{trilyon} & 5th & \tiipa{ሓሙሻይ}{ħamushay} \tiipa{ሓሙሸይቲ}{ħamushäyti} \\
6 & \tiipa{ሽዱሽተ}{shdushtä} & 60 & \tiipa{ሱሳ}{susa} & $10^{15}$ & \tiipa{ኳድሪልዮን}{kwadrilyon} & 6th & \tiipa{ሻድሻይ}{shadshay} \tiipa{ሻድሸይቲ}{shadshäyti} \\
7 & \tiipa{ሸውዓተ}{shäw`atä} & 70 & \tiipa{ሰብዓ}{säb`a} & $10^{18}$ & \tiipa{ኵንቲልዮን}{kwntilyon} & 7th\footnotemark & \tiipa{ሻውዓይ}{shaw`ay} \tiipa{ሻውዐይቲ}{shaw`äyti} \\
8 & \tiipa{ሸሞንተ}{shämontä} & 80 & \tiipa{ሰማንያ}{sämanya} & $10^{21}$ & \tiipa{ሰክስቲልዮን}{säkstilyon} & 8th & \tiipa{ሻምናይ}{shamnay} \tiipa{ሻምነይቲ}{shamnäyti} \\
9 & \tiipa{ትሽዓተ}{tsh`atä} & 90 & \tiipa{ቴስዓ}{tes`a} & $10^{24}$ & \tiipa{ሰፕቲልዮን}{säptilyon} & 9th & \tiipa{ታሽዓይ}{tash`ay} \tiipa{ታሽዐይቲ}{tash`äyti} \\
 & & & & & & 10th & \tiipa{ዓስራይ}{`asray} \tiipa{ዓስረይቲ}{`asräyti} \\
\bottomrule
\end{tabular}
\end{table}
\footnotetext{The ordinal 7\textsuperscript{th} is also read as \tiipa{ሻብዓይ}{shab`ay} or \tiipa{ሻብዐይቲ}{shab`äyti}, but less frequently than \tiipa{ሻውዓይ}{shaw`ay}; \tiipa{ሻውዐይቲ}{shaw`äyti}.}

\begin{table}[t]
\centering
\caption{Tigrinya Month Names}
\label{tab:months}
\footnotesize
\begin{tabular}{@{\hspace{3pt}}clclclcl@{\hspace{3pt}}}
\toprule
1 & \tiipa{ጥሪ}{Tri} & 2 & \tiipa{ለካቲት}{läkatit} & 3 & \tiipa{መጋቢት}{mägabit} & 4 & \tiipa{ሚያዝያ}{miyazya} \\
5 & \tiipa{ግንቦት}{gnbot} & 6 & \tiipa{ሰነ}{sänä} & 7 & \tiipa{ሓምለ}{ħamlä} & 8 & \tiipa{ነሓሰ}{näħasä} \\
9 & \tiipa{መስከረም}{mäskäräm} & 10 & \tiipa{ጥቅምቲ}{Tqmti} & 11 & \tiipa{ሕዳር}{ħdar} & 12 & \tiipa{ታሕሳስ}{taħsas} \\
\bottomrule
\end{tabular}
\end{table}

\section{Verbalization Rules}

The core complexity in Tigrinya number verbalization lies in the conjunction system and the determination of compound versus simple structures.

\vspace{-0.8em}
\subsection{The Conjunction System}
\label{sec:conjunction_system}

The suffix \tiipa{ን}{n} functions as a conjunction meaning `and.' It attaches to components of compound numbers to link them. The rules governing its application are:

\vspace{-0.8em}
\begin{enumerate}
 \item \textbf{Single digit numbers}: No conjunction, e.g., 5 $\rightarrow$ \tiipa{ሓሙሽተ}{ħamushtä}

 \item \textbf{Compound numbers}: Conjunction suffix attached to each component, e.g., 25 $\rightarrow$ \ti{ዕስራ}\tib{ን}\ti{ ሓሙሽተ}\tib{ን}\tiipa{}{`sran ħamushtän} 

 \item \textbf{Teen exception}: Numbers 11--19 do not take internal conjunction but do receive the final conjunction when part of a larger compound.
 Example: 15 $\rightarrow$ \tiipa{ዓሰርተ ሓሙሽተ}{`asärtä ħamushtä}
\end{enumerate}

\vspace{-1.2em}
\subsection{Hundred Form Alternation}
\label{sec:hundred_form}

The word for hundred exhibits allomorphic variation: \tiipa{ሚእቲ}{mi'ti} is used in standalone contexts (e.g., 200 $\rightarrow$ \tiipa{ክልተ ሚእቲ}{kltä mi'ti}), while \ti{ሚእ}\tib{ት}\ti{ን}\tiipa{}{mi'tn} is used in compounds, i.e., \tiipa{ት}{t} replaces \tiipa{ቲ}{ti} and is followed by the conjunction suffix, so 203 is read as \tiipa{ክልተ ሚእትን ሰለስተን}{kltä mi'tn sälästän}.

\subsection{Conjunction on Scale Words}
\label{sec:scale_conjunction}

Scale words (thousand, million, etc.) follow the core principle as other number components: they receive the conjunction suffix \tiipa{ን}{n} when part of a \emph{compound} expression. A scale is considered standalone only when it represents the entire number at that magnitude; otherwise, it is part of a multi-component compound and carries conjunction.
This parallels the hundred alternation logic (\S\ref{sec:hundred_form}) but without lexical change of the root words. This is an important structural pattern for fluent verbalization of Tigrinya numbers that is not explicitly documented in the literature.

\vspace{-0.9em}
\begin{itemize}
 \item 25,000 $\rightarrow$ \tiipa{ዕስራን ሓሙሽተን ሽሕ}{`sran ħamushtän shħ} \hfill (single scale level $\Rightarrow$ standalone)
 \item 25,001 $\rightarrow$ \ti{ዕስራን ሓሙሽተን ሽሕ}\tib{ን}\ti{ ሓደ}\tib{ን}\tiipa{}{`sran ħamushtän shħn ħadän} \hfill (scale + units $\Rightarrow$ compound)
 \item 1,025,000 $\rightarrow$ \ti{ሓደ ሚልዮን}\tib{ን} \ti{ዕስራን ሓሙሽተን ሽሕ}\tib{ን} \hfill (millions + thousands $\Rightarrow$ multiscale compound) \\
\hspace*{1.9cm}\tiipa{}{ħadä miljonn `isran ħamushtän shħn}
\end{itemize}

\subsection{Dates, Times, and Currency}

\paragraph{Dates.} Tigrinya dates follow a month-day ordering using Gregorian month names (Table~\ref{tab:months}). Days and years are expressed as cardinal numbers with implicit conjunction for compounds, while months can be read either by name or as a number.

Example: December 25 $\rightarrow$ \tiipa{ታሕሳስ ዕስራን ሓሙሽተን}{taħsas `sran ħamushtän}

\paragraph{Times.} Time expressions place \tiipa{ሰዓት}{sä`at} (hour) first, followed by the hour value and optionally minutes marked with \tiipa{ደቒቅ}{däQiq} (minute) and seconds marked with \tiipa{ካልኢት}{kal'it} (second). Simple minute values receive the conjunction suffix only when the minute marker is omitted; otherwise, the marker itself carries the conjunction.

Example: 3:30 $\rightarrow$ \ti{ሰዓት ሰለስተን ሰላሳ}\tib{ን}\tiipa{}{sä`at sälästän sälasan} \hfill (without minute marker, hence conjunction on unit) \\
\hspace*{2.65cm}\ti{ሰዓት ሰለስተን ሰላሳ ደቒቅ}\tib{ን}\tiipa{}{sä`at sälästän sälasa däk'ik'n} \hfill (conjunction carried by minute marker)

\paragraph{Currency.} Currency expressions apply cardinal rules to numeric amounts, with conjunction suffixes on the currency and subunit (e.g., \tiipa{ሳንቲም}{santim}) to link components.

Example: 5.55 ERN $\rightarrow$ \ti{ሓሙሽተ ናቕፋ}\tib{ን} \ti{ሓምሳን ሓሙሽተን ሳንቲም}\tib{ን} \tiipa{}{ħamushtä nak'fanħamsan ħamushtän santimn}

\paragraph{Telephone Numbers.} Phone numbers are commonly read in digit pairs or single digits. Pairs beginning with zero are read digit-by-digit; others are read as two-digit numbers.

Example: 07123456 $\rightarrow$ \ti{ዜሮ ሸውዓተ ዓሰርተ ክልተ ሰላሳን ኣርባዕተን ሓምሳን ሽዱሽተን} \\
\hspace*{3.4cm}\tiipa{}{zero shäw`atä `asärtä kltä sälasan 'arba`tän ħamsan shdushtän}

\section{Algorithm}

We formalize the cardinal number verbalization as Algorithm~\ref{alg:cardinal}. The key insight is the decomposition into ``parts'', units that receive the conjunction suffix in compound contexts.
The algorithm focuses on cardinal numbers as the core building block, and the extensions for negative numbers, decimals, ordinals, and other classes build on it naturally. For instance, as described in \S\ref{sec:cardinals}, negatives are handled by prefixing the cardinal reading with \tiipa{ኣሉታ}{'aluta}, while decimals are verbalized by converting the integer part, appending \tiipa{ነጥቢ}{näTbi} (point), and reading each digit in the mantissa individually.

\begin{algorithm}[t]
\caption{Cardinal Number Verbalization}
\label{alg:cardinal}
\begin{algorithmic}[1]
\REQUIRE Integer $n \geq 0$
\ENSURE Tigrinya word representation
\IF{$n = 0$}
 \STATE \textbf{return} \tiipa{ዜሮ}{zero}
\ENDIF
\STATE $\text{parts} \gets []$
\FOR{each scale $(v, w)$ in $[(10^{21}, \text{\tiipa{ሰክስቲልዮን}{säkstilyon}}), \ldots, (10^3, \text{\tiipa{ሽሕ}{shħ}})]$}
 \IF{$n \geq v$}
 \STATE $m \gets \lfloor n / v \rfloor$; \quad $n \gets n \mod v$
 \IF{$\textsc{IsSimple}(m)$}
 \STATE Append $\textsc{Convert}_{<1000}(m) + \text{`` ''} + w$ as single part
 \ELSE
 \STATE Append all parts from $\textsc{Convert}_{<1000}(m)$
 \STATE Append $w$ as separate part
 \ENDIF
 \ENDIF
\ENDFOR
\IF{$n > 0$}
 \STATE Append parts from $\textsc{Convert}_{<1000}(n)$
\ENDIF
\IF{$|\text{parts}| = 1$}
 \STATE \textbf{return} $\text{parts}[0]$ with \tiipa{ሚእት}{mi't} $\rightarrow$ \tiipa{ሚእቲ}{mi'ti}
\ELSE
 \STATE \textbf{return} Join parts with conjunction \tiipa{ን}{n} suffix on each
\ENDIF
\end{algorithmic}
\end{algorithm}

In the algorithm, the predicate $\textsc{IsSimple}(m)$ returns true iff $m \in \{1, \ldots, 19\} \cup \{20, 30, \ldots, 90\} \cup \{100, 200, \ldots, 900\}$.

\subsection{Implementation}

We release an open-source implementation of the algorithm and above discussed rules, covering seven categories: cardinals, ordinals, percentages, currency, dates, times, and phone numbers. The package provides entry functions for each category that include optional flags to control alternative forms such as currency names, use \ti{ዜሮ} vs. \ti{ባዶ}, whether to read phone numbers as pairs or as individual digits, etc. Unit tests cover cardinals (0 to $10^{24}$), ordinals, currency with multiple denominations, date/time edge cases, and phone number formatting.

\vspace{1cm}

The main functions in the Tigrinya Numbers package are:

\begin{itemize}[noitemsep, topsep=0pt]
 \item \texttt{num\_to\_cardinal(n, feminine=[T/F])}: Cardinal numbers, negatives, and decimals
 \item \texttt{num\_to\_ordinal(n, feminine=[T/F])}: Ordinal numbers verbalization
 \item \texttt{num\_to\_currency(amount, currency)}: Currency verbalization for a given denomination
 \item \texttt{num\_to\_date(day, month, year)}: Date verbalization with optional parameters
 \item \texttt{num\_to\_time(hour, minute, second)}: Time verbalization with optional parameters
 \item \texttt{num\_to\_phone(phone\_str, use\_singles=[T/F])}: Phone number verbalization in pairs or single digits
 \item \texttt{num\_to\_percent(n)}: Percentage verbalization, adds the suffix \tiipa{ሚእታዊት}{mi'tawit} to cardinal reading
\end{itemize}

\section{Evaluation of Large Language Models}

To assess whether current large language models (LLMs) have internalized Tigrinya number verbalization rules, we constructed an evaluation set with 100 examples spanning six categories: cardinals (50), ordinals (15), currency (10), dates (10), times (10), and phone numbers (5). The set emphasizes challenging cases: compound numbers requiring conjunction placement, teens that break the standard pattern, scale words with compound multipliers, and suppletive ordinal forms. We diversified digit usage beyond common values to test true linguistic competence rather than memorized patterns.

\begin{figure}[t]
\centering
\includegraphics[width=0.975\columnwidth]{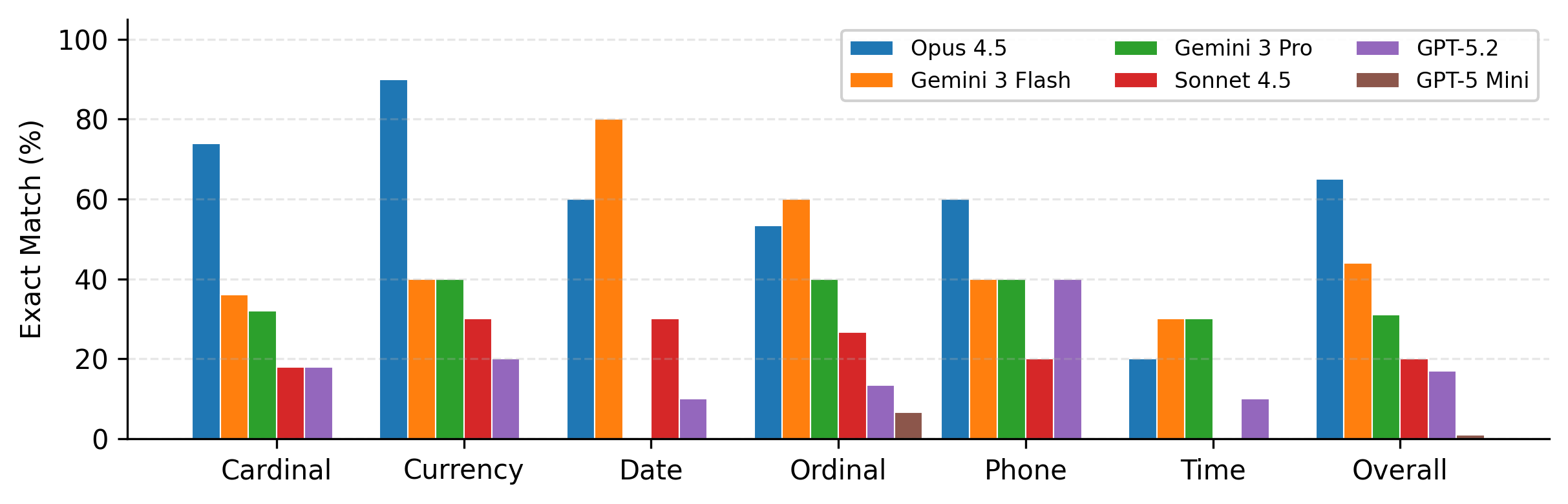}
\caption{LLM performance comparison across categories. Best overall: Opus 4.5 (65\%), followed by Gemini 3 Flash (44\%).}
\label{fig:llm_results}
\end{figure}

We evaluated six frontier models from three major providers. Each model was prompted with the verbalization task including the category name as context. Accuracy was measured via exact string match after Unicode normalization. The results (Table~\ref{tab:llm_eval}, Figure~\ref{fig:llm_results}) reveal substantial deficiencies. While models achieve moderate accuracy on simple cardinals and currency, performance degrades significantly for other categories. Performance depends on the models' familiarity with Tigrinya. Strikingly, GPT-5 Mini struggled to give correct results in almost all cases within two token budget settings (2048 and 4096) per request. Common errors include: (1) partial answers that include digits, typographical errors in the base words, and loanwords from related languages such as Amharic; (2) omitting the necessary conjunction suffix; (3) incorrect application to teens; and (4) failure to distinguish simple vs. compound multipliers with scale words. These findings underscore the value of explicit rule documentation and deterministic implementations for production NLP systems.

\begin{table}[t]
\centering
\caption{Performance of LLMs on Tigrinya Number Verbalization. GPT-5 Mini runs out of max tokens (2048 \& 4096) for most requests.}
\label{tab:llm_eval}
\begin{tabular}{@{\hspace{7pt}}lrrrrrrr@{\hspace{7pt}}}
\toprule
\textbf{Model} & \textbf{Cardinal} & \textbf{Currency} & \textbf{Date} & \textbf{Ordinal} & \textbf{Phone} & \textbf{Time} & \textbf{Overall (\%)} \\
\midrule
 Gemini 3 Flash & 18/50 & 4/10 & \textbf{8/10} & \textbf{9/15} & 2/5 & \textbf{3/10} & 44 \\
 Gemini 3 Pro & 16/50 & 4/10 & 0/10 & 6/15 & 2/5 & 3/10 & 31 \\
 GPT-5 Mini & 0/50 & 0/10 & 0/10 & 1/15 & 0/5 & 0/10 & 1 \\
 GPT-5.2 & 9/50 & 2/10 & 1/10 & 2/15 & 2/5 & 1/10 & 17 \\
 Opus 4.5 & \textbf{37/50} & \textbf{9/10} & 6/10 & 8/15 & \textbf{3/5} & 2/10 & \textbf{65} \\
 Sonnet 4.5 & 9/50 & 3/10 & 3/10 & 4/15 & 1/5 & 0/10 & 20 \\
\bottomrule
\end{tabular}
\end{table}

\section{Applications}

The formalized rules and implementation address several practical needs:
\textbf{Text-to-Speech (TTS)} synthesis requires text normalization that converts digits to pronounceable words. Prior work on Tigrinya TTS \citep{keletay-worku-2020-tts, pratap-etal-2023-mms-tts-tir, mihreteab-etal-2025-end-to-end-tts} has not addressed systematic number handling. Our implementation provides a drop-in preprocessing component.
\textbf{Automatic Speech Recognition (ASR)} language models benefit from expanded text corpora that include number words. The deterministic nature of our rules enables systematic generation of training data (augmentation) for \textbf{Language Modeling}. Similarly, the rules documented in this work can serve as structured knowledge for fine-tuning or prompting LLMs to improve their overall Tigrinya capabilities.
\textbf{Assistive Technologies} such as screen readers for users with visual impairments require accurate number verbalization for document reading, form filling, and general accessibility.

\section{Conclusion}

This work provides the first systematic formalization of Tigrinya number verbalization rules, addressing an underserved area in computational linguistics for Semitic languages of the Horn of Africa. The conjunction-based compound structure, the simple/compound scale word distinction, and the hundred allomorphy represent linguistic patterns that require explicit documentation for computational implementation.
The released implementation and test suite establish a foundation for Tigrinya NLP applications requiring number handling. More broadly, this work contributes to language preservation, accessibility for speakers with disabilities, and reduced technological disparity for low-resource language communities. Future work includes extending coverage to mathematical expressions, as well as integration with the broader ecosystem.

\textbf{Limitations:} (1) The LLM evaluation assumes basic support for Tigrinya by the models, but it should be noted that the model providers do not officially claim to support Tigrinya. The evaluations are indicative using a limited test set and should be expanded in future work as the models improve. (2) There are regional dialects of Tigrinya in Eritrea and Ethiopia with minor orthographic and spelling variations, when in doubt the implementation in this work defaults to the Eritrean variant but it can be extended to others with minor modifications.

\bibliography{references}
\bibliographystyle{icml2025}

\newpage
\appendix
\section{Evaluation Set}
\label{appendix:eval}

\footnotesize
\begin{longtable}{@{\hspace{1pt}}llp{5.65cm}|llp{5.65cm}@{\hspace{1pt}}}
\caption{The evaluation set used for LLM assessment. Entries shows the input and ground truth answer(s), separated by semicolons.}
\label{tab:eval_set_samples} \\
\toprule
\textbf{\#} & \textbf{Input} & \textbf{Ground Truth} & \textbf{\#} & \textbf{Input} & \textbf{Ground Truth} \\
\midrule
\endhead
\multicolumn{6}{c}{\textbf{Cardinal Number Evaluation Examples (50 entries)}} \\
\addlinespace
1 & 84{,}001 & \ti{ሰማንያን ኣርባዕተን ሽሕን ሓደን} & 26 & 987 & \ti{ትሽዓተ ሚእትን ሰማንያን ሸውዓተን} \\
2 & 1{,}234{,}567 & \ti{ሓደ ሚልዮንን ክልተ ሚእትን ሰላሳን ኣርባዕተን ሽሕን ሓሙሽተ ሚእትን ሱሳን ሸውዓተን} & 27 & 12{,}345 & \ti{ዓሰርተ ክልተ ሽሕን ሰለስተ ሚእትን ኣርብዓን ሓሙሽተን} \\
3 & 147{,}001 & \ti{ሚእትን ኣርብዓን ሸውዓተን ሽሕን ሓደን} & 28 & 111{,}111 & \ti{ሚእትን ዓሰርተ ሓደን ሽሕን ሚእትን ዓሰርተ ሓደን} \\
4 & 40 & \ti{ኣርብዓ} & 29 & 5{,}555 & \ti{ሓሙሽተ ሽሕን ሓሙሽተ ሚእትን ሓምሳን ሓሙሽተን} \\
5 & 34{,}700 & \ti{ሰላሳን ኣርባዕተን ሽሕን ሸውዓተ ሚእትን} & 30 & 37{,}000 & \ti{ሰላሳን ሸውዓተን ሽሕ} \\
6 & 101{,}000 & \ti{ሓደ ሚእትን ሓደን ሽሕ; ሚእትን ሓደን ሽሕ} & 31 & 84{,}000 & \ti{ሰማንያን ኣርባዕተን ሽሕ} \\
7 & 14 & \ti{ዓሰርተ ኣርባዕተ} & 32 & 147{,}000 & \ti{ሚእትን ኣርብዓን ሸውዓተን ሽሕ} \\
8 & 17 & \ti{ዓሰርተ ሸውዓተ} & 33 & 3{,}007 & \ti{ሰለስተ ሽሕን ሸውዓተን} \\
9 & 18 & \ti{ዓሰርተ ሸሞንተ} & 34 & 4{,}019 & \ti{ኣርባዕተ ሽሕን ዓሰርተ ትሽዓተን} \\
10 & 19 & \ti{ዓሰርተ ትሽዓተ} & 35 & 7{,}348 & \ti{ሸውዓተ ሽሕን ሰለስተ ሚእትን ኣርብዓን ሸሞንተን} \\
11 & 23 & \ti{ዕስራን ሰለስተን} & 36 & 9{,}876 & \ti{ትሽዓተ ሽሕን ሸሞንተ ሚእትን ሰብዓን ሽዱሽተን} \\
12 & 37 & \ti{ሰላሳን ሸውዓተን} & 37 & 1{,}001{,}000 & \ti{ሓደ ሚልዮንን ሓደ ሽሕን; ሚልዮንን ሽሕን} \\
13 & 48 & \ti{ኣርብዓን ሸሞንተን} & 38 & 4{,}000{,}003 & \ti{ኣርባዕተ ሚልዮንን ሰለስተን} \\
14 & 69 & \ti{ሱሳን ትሽዓተን} & 39 & 7{,}894{,}321 & \ti{ሸውዓተ ሚልዮንን ሸሞንተ ሚእትን ቴስዓን ኣርባዕተን ሽሕን ሰለስተ ሚእትን ዕስራን ሓደን} \\
15 & 84 & \ti{ሰማንያን ኣርባዕተን} & 40 & 37{,}000{,}000 & \ti{ሰላሳን ሸውዓተን ሚልዮን} \\
16 & 93 & \ti{ቴስዓን ሰለስተን} & 41 & 1{,}000{,}001 & \ti{ሓደ ሚልዮንን ሓደን; ሚልዮንን ሓደን} \\
17 & 25{,}000 & \ti{ዕስራን ሓሙሽተን ሽሕ} & 42 & -7 & \ti{ኣሉታ ሸውዓተ} \\
18 & 700 & \ti{ሸውዓተ ሚእቲ} & 43 & -38 & \ti{ኣሉታ ሰላሳን ሸሞንተን} \\
19 & 103 & \ti{ሓደ ሚእትን ሰለስተን; ሚእትን ሰለስተን} & 44 & -749 & \ti{ኣሉታ ሸውዓተ ሚእትን ኣርብዓን ትሽዓተን} \\
20 & 118 & \ti{ሚእትን ዓሰርተ ሸሞንተን} & 45 & 3.14 & \ti{ሰለስተ ነጥቢ ሓደ ኣርባዕተ} \\
21 & 147 & \ti{ሚእትን ኣርብዓን ሸውዓተን} & 46 & 0.7 & \ti{ዜሮ ነጥቢ ሸውዓተ} \\
22 & 309 & \ti{ሰለስተ ሚእትን ትሽዓተን} & 47 & 8.03 & \ti{ሸሞንተ ነጥቢ ዜሮ ሰለስተ} \\
23 & 438 & \ti{ኣርባዕተ ሚእትን ሰላሳን ሸሞንተን} & 48 & 47.893 & \ti{ኣርብዓን ሸውዓተን ነጥቢ ሸሞንተ ትሽዓተ ሰለስተ} \\
24 & 674 & \ti{ሽዱሽተ ሚእትን ሰብዓን ኣርባዕተን} & 49 & 123.007 & \ti{ሓደ ሚእትን ዕስራን ሰለስተን ነጥቢ ዜሮ ዜሮ ሸውዓተ} \\
25 & 819 & \ti{ሸሞንተ ሚእትን ዓሰርተ ትሽዓተን} & 50 & 99 & \ti{ቴስዓን ትሽዓተን} \\
\midrule
\multicolumn{6}{c}{\textbf{Ordinal Number Evaluation Examples (15 entries)}} \\
\addlinespace
51 & 1st (M) & \ti{ቀዳማይ} & 59 & 8th (F) & \ti{ሻምነይቲ} \\
52 & 3rd (M) & \ti{ሳልሳይ} & 60 & 9th (F) & \ti{ታሽዐይቲ} \\
53 & 7th (M) & \ti{ሻውዓይ} & 61 & 13th & \ti{መበል ዓሰርተ ሰለስተ} \\
54 & 8th (M) & \ti{ሻምናይ} & 62 & 17th & \ti{መበል ዓሰርተ ሸውዓተ} \\
55 & 9th (M) & \ti{ታሽዓይ} & 63 & 38th & \ti{መበል ሰላሳን ሸሞንተን} \\
56 & 1st (F) & \ti{ቀዳመይቲ} & 64 & 74th & \ti{መበል ሰብዓን ኣርባዕተን} \\
57 & 4th (F) & \ti{ራብዐይቲ} & 65 & 147th & \ti{መበል ሓደ ሚእትን ኣርብዓን ሸውዓተን} \\
58 & 6th (F) & \ti{ሻድሸይቲ} & & & \\
\midrule
\multicolumn{6}{c}{\textbf{Currency Evaluation Examples (10 entries)}} \\
\addlinespace
66 & 7 ERN & \ti{ሸውዓተ ናቕፋ} & 71 & 83.09 ERN & \ti{ሰማንያን ሰለስተን ናቕፋን ትሽዓተ ሳንቲምን} \\
67 & 300 ERN & \ti{ሰለስተ ሚእቲ ናቕፋ} & 72 & 347.68 ERN & \ti{ሰለስተ ሚእትን ኣርብዓን ሸውዓተን ናቕፋን ሱሳን ሸሞንተን ሳንቲምን} \\
68 & 4{,}789 ERN & \ti{ኣርባዕተ ሽሕን ሸውዓተ ሚእትን ሰማንያን ትሽዓተን ናቕፋ} & 73 & 0.37 ERN & \ti{ሰላሳን ሸውዓተን ሳንቲም} \\
69 & 7.43 ERN & \ti{ሸውዓተ ናቕፋን ኣርብዓን ሰለስተን ሳንቲምን} & 74 & 0.89 ERN & \ti{ሰማንያን ትሽዓተን ሳንቲም} \\
70 & 18.75 ERN & \ti{ዓሰርተ ሸሞንተ ናቕፋን ሰብዓን ሓሙሽተን ሳንቲምን} & 75 & 73 ETB & \ti{ሰብዓን ሰለስተን ብር} \\
\midrule
\multicolumn{6}{c}{\textbf{Date Evaluation Examples (10 entries)}} \\
\addlinespace
76 & 7/3 & \ti{መጋቢት ሸውዓተ; ዕለት ሸውዓተ ወርሒ ሰለስተ} & 81 & 31/7 & \ti{ሓምለ ሰላሳን ሓደን; ዕለት ሰላሳን ሓደን ወርሒ ሸውዓተ} \\
77 & 14/9 & \ti{መስከረም ዓሰርተ ኣርባዕተ; \newline ዕለት ዓሰርተ ኣርባዕተ ወርሒ ትሽዓተ} & 82 & 29/12 & \ti{ታሕሳስ ዕስራን ትሽዓተን; \newline ዕለት ዕስራን ትሽዓተን ወርሒ ዓሰርተ ክልተ} \\
78 & 10/11 & \ti{ዕለት ዓሰርተ ወርሒ ዓሰርተ ሓደ; \newline ሕዳር ዓሰርተ} & 83 & 24/5/1991 & \ti{ዕለት ዕስራን ኣርባዕተን ወርሒ ሓሙሽተ ሽሕን ትሽዓተ ሚእትን ቴስዓን ሓደን; ግንቦት ዕስራን ኣርባዕተን ሽሕን ትሽዓተ ሚእትን ቴስዓን ሓደን} \\
79 & 27/8 & \ti{ነሓሰ ዕስራን ሸውዓተን; \newline ዕለት ዕስራን ሸውዓተን ወርሒ ሸሞንተ} & 84 & 1/9/2023 & \ti{መስከረም ሓደ ክልተ ሽሕን ዕስራን ሰለስተን; ዕለት ሓደ ወርሒ ትሽዓተ ክልተ ሽሕን ዕስራን ሰለስተን} \\
80 & 23/4 & \ti{ሚያዝያ ዕስራን ሰለስተን; \newline ዕለት ዕስራን ሰለስተን ወርሒ ኣርባዕተ} & 85 & 17/2/2007 & \ti{ለካቲት ዓሰርተ ሸውዓተ ክልተ ሽሕን ሸውዓተን; ዕለት ዓሰርተ ሸውዓተ ወርሒ ክልተ ክልተ ሽሕን ሸውዓተን} \\
\midrule
\multicolumn{6}{c}{\textbf{Time Evaluation Examples (10 entries)}} \\
\addlinespace
86 & 3:00 & \ti{ሰዓት ሰለስተ} & 91 & 3:47 & \ti{ሰለስተን ኣርብዓን ሸውዓተን; \newline ሰዓት ሰለስተን ኣርብዓን ሸውዓተን ደቒቕን} \\
87 & 9:00 & \ti{ሰዓት ትሽዓተ} & 92 & 9:38 & \ti{ሰዓት ትሽዓተን ሰላሳን ሸሞንተን; ሰዓት ትሽዓተን ሰላሳን ሸሞንተን ደቒቕን; ትሽዓተን ሰላሳን ሸሞንተን} \\
88 & 7:30 & \ti{ሰዓት ሸውዓተን ሰላሳ ደቒቕን; \newline ሰዓት ሸውዓተን ሰላሳን} & 93 & 11:54 & \ti{ዓሰርተ ሓደን ሓምሳን ኣርባዕተን; \newline ሰዓት ዓሰርተ ሓደን ሓምሳን ኣርባዕተን ደቒቕን} \\
89 & 4:15 & \ti{ሰዓት ኣርባዕተን ዓሰርተ ሓሙሽተ ደቒቕን; \newline ሰዓት ኣርባዕተን ዓሰርተ ሓሙሽተን} & 94 & 2:37:48 & \ti{ሰዓት ክልተን ሰላሳን ሸውዓተን ደቒቕን ኣርብዓን ሸሞንተን ካልኢትን} \\
90 & 8:10 & \ti{ሰዓት ሸሞንተን ዓሰርተ ደቒቕን; ሰዓት ሸሞንተን ዓሰርተን; ሸሞንተን ዓሰርተ ደቒቕን; ሸሞንተን ዓሰርተን} & 95 & 6:14:29 & \ti{ሰዓት ሽዱሽተን ዓሰርተ ኣርባዕተ ደቒቕን ዕስራን ትሽዓተን ካልኢትን} \\
\midrule
\multicolumn{6}{c}{\textbf{Phone Number Evaluation Examples (5 entries)}} \\
\addlinespace
96 & 07-34-89 & \ti{ዜሮ ሸውዓተ ሰለስተ ኣርባዕተ ሸሞንተ ትሽዓተ; \newline ዜሮ ሸውዓተ ሰላሳን ኣርባዕተን ሰማንያን ትሽዓተን} & 99 & 83-47-19 & \ti{ሰማንያን ሰለስተን ኣርብዓን ሸውዓተን ዓሰርተ ትሽዓተ; ሸሞንተ ሰለስተ ኣርባዕተ ሸውዓተ ሓደ ትሽዓተ} \\
97 & 01-78-43 & \ti{ዜሮ ሓደ ሰብዓን ሸሞንተን ኣርብዓን ሰለስተን; \newline ዜሮ ሓደ ሸውዓተ ሸሞንተ ኣርባዕተ ሰለስተ} & 100 & 07-18-43-97 & \ti{ዜሮ ሸውዓተ ዓሰርተ ሸሞንተ ኣርብዓን ሰለስተን ቴስዓን ሸውዓተን; ዜሮ ሸውዓተ ሓደ ሸሞንተ ኣርባዕተ ሰለስተ ትሽዓተ ሸውዓተ} \\
98 & 17-38-94 & \ti{ዓሰርተ ሸውዓተ ሰላሳን ሸሞንተን ቴስዓን ኣርባዕተን} & & & \\
\bottomrule
\end{longtable}
 
\end{document}